\documentclass[letterpaper, 10 pt, conference]{ieeeconf}  % Comment this line out if you need a4paper

\usepackage{graphicx}% http://ctan.org/pkg/graphicx
\usepackage{lipsum}% http://ctan.org/pkg/lipsum
\usepackage{xcolor}
\usepackage{hyperref}

\IEEEoverridecommandlockouts                              % This command is only needed if 
\title{\LARGE \bf
Simple Kinematic Feedback Enhances Autonomous Learning in Bio-Inspired Tendon-Driven Systems
}

\author{Ali Marjaninejad$^{1}$, Darío Urbina-Meléndez$^{2}$, and Francisco Valero-Cuevas$^{3}$, \it{Senior Member, IEEE}% <-this % stops a space
\thanks{$^{1}$A. Marjaninejad is with University of Southern California, Los Angeles, Ca 90089 USA
        {\tt\small e-mail: marjanin@usc.edu}}%
\thanks{$^{2}$D. Urbina-Meléndez is with University of Southern California, Los Angeles, Ca 90089 USA.
        {\tt\small e-mail: urbiname@usc.edu}}%
\thanks{$^{3}$F. Valero-Cuevas is with University of Southern California, Los Angeles, Ca 90089 USA
        {\tt\small (corresponding author) email-: valero@usc.edu; phone: 213-740-4219}}%
}

\begin{document}

\maketitle
\thispagestyle{empty}
\pagestyle{empty}

%%%%%%%%%%%%%%%%%%%%%%%%%%%%%%%%%%%%%%%%%%%%%%%%%%%%%%%%%%%%%%%%%%%%%%%%%%%%%%%%
\begin{abstract}

Error feedback is known to improve performance by correcting control signals in response to perturbations.
Here we show how adding simple error feedback can also accelerate and robustify autonomous learning in a tendon-driven robot.
We implemented two versions of the General-to-Particular (G2P) autonomous learning algorithm to produce multiple movement tasks using a tendon-driven leg with two joints and three tendons: one with and one without kinematic feedback.
As expected, feedback improved performance in simulation and hardware. However, we see these improvements even in the presence of sensory delays of up to 100 ms and when experiencing substantial contact collisions.
Importantly, feedback accelerates learning and enhances G2P's continual refinement of the initial inverse map by providing the system with more relevant data to train on. This allows the system to perform well even after only 60 seconds of initial motor babbling.

\end{abstract}

%%%%%%%%%%%%%%%%%%%%%%%%%%%%%%%%%%%%%%%%%%%%%%%%%%%%%%%%%%%%%%%%%%%%%%%%%%%%%%%%
\section{INTRODUCTION}

\begin{figure*}
    \centering
    \includegraphics[width=.9\linewidth]{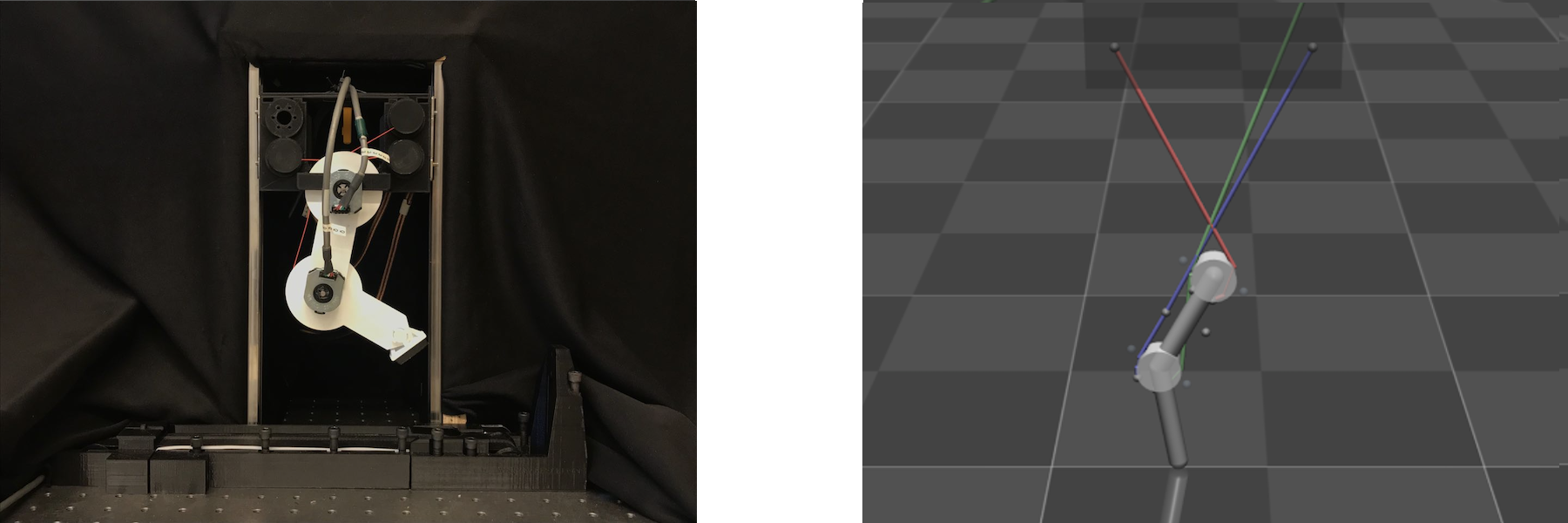}
    \caption{Physical tendon-driven robotic limb (left) and the simulated system in MuJoCo environment (Right)}
    \label{fig:Fig1}
\end{figure*}

\begin{figure}
\centering
\includegraphics[width=\linewidth]{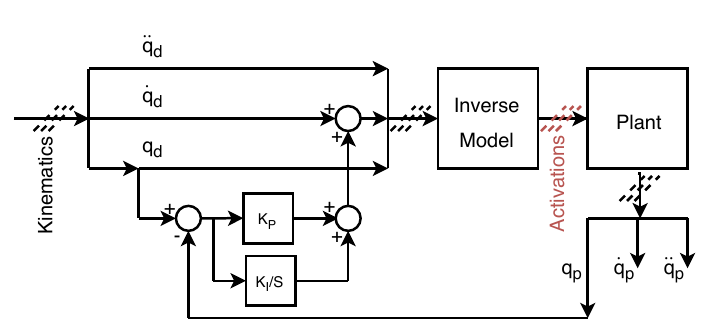}
\caption{Schematic of the close-loop system.}
\label{fig:Fig2}
\end{figure}

The field of robotics in general would benefit greatly from autonomous learning to control movements with minimal prior knowledge and limited experience \cite{marjaninejad2019autonomous,kwiatkowski2019task,nagabandi2018neural}.
Extensive trial-and-error experience in the real world can be very costly both in biological and robotic systems. Not only does it risk injury, but the opportunity cost can be large. Therefore, evolutionary game theory in biological systems favors systems that can function suboptimaly or well-enough with only limited experience, and continue to learn on-the-go from every experience~\cite{marjaninejad2019autonomous}. Biological systems can then use sensory feedback to refine performance as needed. 

Such learning from limited experience is also attractive in robotics~\cite{kwiatkowski2019task,marjaninejad2019autonomous}, mostly in situations where optimality is not as critical as adaptability to unstructured environments, unpredictable payloads, or working with systems for which creating accurate models is costly or time consuming. Thus data-efficient learning that produces suboptimal behavior can be a practical and attractive control strategy as it does not rely on accurate prior models or extensive expert knowledge ~\cite{kobayashi2003adaptive, marques2014spontaneous, morimoto2001acquisition, gijsberts2013real, geijtenbeek2013flexible, rombokas2012tendon, hunt2017development, kumar2014real}
, or require thousands of hours of or learning in simulation~\cite{mnih2015human, takahashi2017dynamic, heess2017emergence, andrychowicz2018learning, schulman2017proximal, schulman2015trust} (please see \cite{marjaninejad2019autonomous} for detailed discussion on how our General-to-Particular (G2P) algorithm relates to the field).

A drawback of learning with limited experience that produces suboptimal behavior is that the performance of the model can degrade when encountering dynamics far from those under which it was trained. On the other hand, systems that heavily depend on the feedback error correction would not perform efficiently and are prone to instability, especially in the presence of sensory delays\cite{bongard2006resilient, marjaninejad2019autonomous}. Moreover, it is important to note that in a tendon-driven system, actuation is not directly connected to the joint and therefore, a simple off the shelf PID controller cannot be used without knowing the dynamical equations of the system~\cite{marjaninejad2019autonomous,valero2016fundamentals,marjaninejad2019should}. Thus, here we explore the combination of data-efficient learning algorithm, G2P, with simple feedback to maintain the key benefits of learning under limited experience while improving performance and robustness to perturbations (or unmodeled dynamics) as needed. This approach is directly inspired by biological systems that, under certain circumstances, successfully use simple corrective responses triggered by delayed and non-collocated sensory signals~\cite{milton2009balancing, cetinkaya2018stabilizing, cetinkaya2015sampled}.

 As an initial proof-of-principle, we implemented two versions of the data-efficient autonomous learning algorithm, G2P~\cite{marjaninejad2019autonomous} (that is originally designed to control tendon-driven systems) : One purely feedforward (open-loop) as published in ~\cite{marjaninejad2019autonomous}, and one with simple feedback on joint angles (close-loop). Both implementations of G2P find motor commands that produce desired leg kinematics by creating an inverse map. The initial inverse map is generated from ``motor babbling'' input-output data (i.e., random sequences of input commands to the three motors driving the tendons that produce time histories of two joint angles of the leg). We find that the performance is, as expected, better for the close-loop system as it compensates for errors in the leg joint angles arising from imperfections of the inverse map or external perturbations (e.g., contact dynamics). However, we also find that, by collecting more task relevant data, this simple feedback accelerates learning and improves the quality of the inverse map that enables the system to work with shorter motor babbling sessions and also improves learning on-the-go capability (the refinement of the inverse map from each experience) of the system. In addition, we also report improved performance even when sensory signals are delayed. We have validated our method on a physical tendon-driven leg to demonstrate its utility in real-world applications.

\section{Methods}
In this section, we first discuss the design of the tendon-driven system. Next, we formulate and describe the controller design. Lastly, we discuss the tests we performed in detail.
\subsection{Tendon-driven leg design}
Tendon-driven anatomies are a relevant use case because they are difficult to control as they are simultaneously nonlinear, under-determined and over-determined~\cite{marjaninejad2019should, valero2016fundamentals, marjaninejad2019autonomous, cohn2018feasibility}. The simulated leg is a similar design to the physical leg used in~\cite{marjaninejad2019autonomous}. It is a 2-DoF planar leg actuated with 3 (minimal number of tendons to control a 2-DoF leg) tendons. Unlike ~\cite{marjaninejad2019autonomous}, the simulation model uses a Hill-type model of skeletal muscle (MuJoCo's built in force-length and force-velocity model~\cite{todorov2012mujoco,valero2016fundamentals} and has moment arms that can bowstring. The physical system used for validation is a replica of the one used in ~\cite{marjaninejad2019autonomous} with an improvement on the data acquisition system (we use PXI system from National Instruments, Austin, TX, USA). Figure~\ref{fig:Fig1} shows the tendon-driven legs for the physical and the simulation systems.

\subsection{Controller design}
Our system takes desired movement kinematics (joint angles for each joint and their first two derivatives; angular velocities and angular accelerations) and outputs activation values that will drive the actuators (skeletal muscles in simulation and electric brushless DC motors connected to the tendons on the physical system) to produce the desired kinematics on the leg.

The feedforward path consists of an inverse mapping that maps the desired kinematics to the activations that will ideally (in the case of a flawless inverse map and without any perturbation) create activations required to replicate the desired kinematics on the plant (Figure~\ref{fig:Fig2}). This inverse map is created by training a Multi-Layer Perceptron (MLP) Artificial Neural Network (ANN) with 15 hidden layer neurons using the data collected during a short phase (5 minutes) of random movements and observing their corresponding kinematics which is called motor babbling (please see ~\cite{marjaninejad2019autonomous} or the supplementary code for more details on the feedforward path). Once the activations are calculated, they will be fed into the plant and the corresponding kinematics will be recorded. These observations can then be used in a feedback loop to compensate for any error in the inverse map or the error caused by external perturbation.

Here we only use joint angles as the sensory feedback. Also, we mainly focus on reducing the error on the joint angles (as opposed to its derivatives) since error in joint angles is less forgiving than joint velocity or acceleration in successful compilation of most day to day tasks either being manipulation, locomotion or other movements; however, if desired, user can substitute the error term with the angular velocity or acceleration or a weighted combination of them.
We know that for a given joint, joint angle and angular velocity are related by equation~\ref{equ:Equation1}.
\begin{equation}
    \Delta q = \dot{q} \; . \;dt
    \label{equ:Equation1}
\end{equation}
where $t$ is time. Therefore, we can compensate the error in position by changing the velocity corresponding to the magnitude and the direction of the error. We implement a PI controller like method where we add an adjustment term to the desired angular velocity of each joint proportional to its current and cumulative error (see discussions for alternative choices). Equations~\ref{equ:Equation2}-~\ref{equ:Equation6} describe relationships between all system variables over a complete loop:
\begin{equation}
    a[n]_{(3\times1)} = ANN(q_{c}[n]_{\:(2\times1)}, \dot{q}_{c}[n]_{\:(2\times1)}, \ddot{q}_{c}[n]_{\:(2\times1)})    \label{equ:Equation2}
\end{equation}
where $a[n]$ is the activation vector at time sample $n$ and $q_c$, $\dot{q_c}$, and $\ddot{q_c}$, are joint angle, control angular velocity and control angular acceleration, respectively. These control kinematics are calculated as follows:
\begin{equation}
    \ddot{q_c} = \ddot{q_d}, \;\;\; q_c = q_d
    \label{equ:Equation3}
\end{equation}
\begin{equation}
    \dot{q_c} = \dot{q_d}+\dot{q_a}
    \label{equ:Equation4}
\end{equation}
\begin{equation}
    \dot{q_a} = K_{P\:(2\times2)} \; q_e + K_{I\:(2\times2)} \int{q_e}\:dt
    \label{equ:Equation5}
\end{equation}
\begin{equation}
    q_e = q_d - q_p
    \label{equ:Equation6}
\end{equation}

where subscripts $d$, $a$, and $e$ stand for desired, adjustment, and error respectively. Also, $K_P$ and $K_I$ are diagonal matrices defining the proportional and integral coefficients for each joint. The complete schematic block diagram of the close-loop system is depicted in Figure~\ref{fig:Fig2}.

\subsection{Studied tasks}
To demonstrate the performance of the proposed method and its capabilities, we have tested it in a number of different cases, each of which demonstrates at least one of its prominent features. Wherever applicable, we have compared the results with the ones produced by the open-loop method used in ~\cite{marjaninejad2019autonomous}.

\subsubsection{Cyclical movements in-air task}
During this task, the leg is suspended in-air (i.e. no contact dynamics/external perturbations involved) and is commanded to perform 50 random cyclical patterns (10 cycle with 2.5 seconds each). Since there are no external perturbations applied to the system in this task, an ideal inverse map should be able to perform it flawlessly. However, the inverse map trained with limited experience is almost always imperfect; during this task we will study the effect of the proposed close-loop system on reducing these imperfections.
These patterns are created by projecting a vector of 10 random values sampled from a uniform distribution ($U(0,1)$) into joint angle space as described in ~\cite{marjaninejad2019autonomous}. In short, each random number defines the normalized radial distance from the center of the joint angle space of one of equally distributed spokes (each 36 degrees apart) and then, these points will be connected and the resulting closed cycle will be filtered to make it smooth (see ~\cite{marjaninejad2019autonomous} for more details).

\subsubsection{Point-to-point movements in-air task}
Unlike the continuous and smooth cyclical task, point-to-point task is consisted of discrete ramp-and-hold movements. Since the inverse map was trained with full kinematics, it would be interesting to study how it performs when the desired task involves maintaining joint angles in specific positions (both angular velocity and accelerations will be equal to zero in all these positions). The point-to-point task is designed to study these cases and involved 50 trials where in each trial the leg is going to be commanded to go to 10 random positions ($U(joint_{min}, joint_{max})$ for each joint) and stay there for a predefined duration (2.5 seconds here).

\subsubsection{Different cycle period durations task}

During the motor babbling, the inverse map is introduced to a very sparse set of samples in the 6D kinematics space~\cite{marjaninejad2019autonomous}. Although it has fully swiped across joint angles for both joints during the motor babbling phase, there are many combinations of these angles with their angular velocities and accelerations that will not be experienced. Here we are going to study the performance of the system for perfectly cyclical movements (sin and cos) over a wide range of cycle periods (1 / cycle frequency) to investigate how well the open-loop system performs in each case and to compare the performance of the proposed feedback controller.

\subsubsection{Performance in the presence of contact dynamics tasks}
Dealing with contact dynamics is a current challenge in robotics~\cite{fazeli2019see, fazeli2017learning}. Therefore, it is important to test the performance of the proposed method in the presence of contact dynamics. We have shown that the open-loop system can perform well when introduced to minor contact dynamics\cite{marjaninejad2019autonomous}; however, the performance of the system has not been studied under the effect of significant contact dynamics caused by the need to push the system forward/backward in the presence of an antagonist force or the need to carry its own weight (note that the system was trained in-air and therefore adding weight will be a major change to its dynamics). Here, we have studied the performance of the system during two tasks both including contact dynamics.

\paragraph{Locomotion with the gantry}
In this task, we have lowered the chassis (so that the leg can touch the floor) and let it move on the x-axis (forward/backward) with friction. Moreover, we have held it up with a spring-damper (build-in features in MuJoCo) so that it can partially compensate for the weight (similar to a gantry). Similar to the ``Cyclical movements in-air'', here we have applied 50 different cyclical movements and studied the performance of the system. Please note that here we are simply applying random cyclical movements to compare open-loop and close-loop performance; however, a higher-level controller can also be used to find better movement trajectories to yield higher forward displacement~\cite{marjaninejad2019autonomous}.

\paragraph{Holding a posture under a weight}
In this task, we took off the spring-damper mechanism provided by the gantry and increased the weight of the chassis significantly. The goal here for the leg is to stay vertically straight (standing leg position) while reacting to a strong downward force applied to it, due to the added weight of the gantry.

\subsubsection{Learning from each experience task}
Experience can be very costly in real-world physical systems~\cite{marjaninejad2019autonomous} and therefore, an efficient system should be able to start performing as soon as possible and improving the performance with the data coming from each experience. During this task, we start with an inverse map created using a shorter duration (1 minute) and run the system on a cyclical task for 25 repetitions; after each repetition, we refine the inverse map with the cumulative data from all the experience that the system had so far (including the motor babbling). We repeat this process for 50 different cyclical trajectories.

\subsubsection{Variable feedback delay task}
Delay in the sensory feedback or processing information is inevitable in real-world applications. In a system that solely depends on error correction, these delays can inject large errors and even drive the system to instability. We have studied the performance of the system over a wide range of loop delays (from 5 to 100 ms; which is about the largest delay in the human sensory-motor loop) over 50 random cyclical movements.

All tasks were performed on both simulation (sim) and physical (phys) systems except "performance in the presence of the contact dynamics" and "variable feedback delay" tasks which were only performed on simulation due to physical limitations. Also, physical results for the "learning from each experience task" has already been studied on \cite{marjaninejad2019autonomous}.

\section{Results}

\begin{figure}
\includegraphics[width=\linewidth]{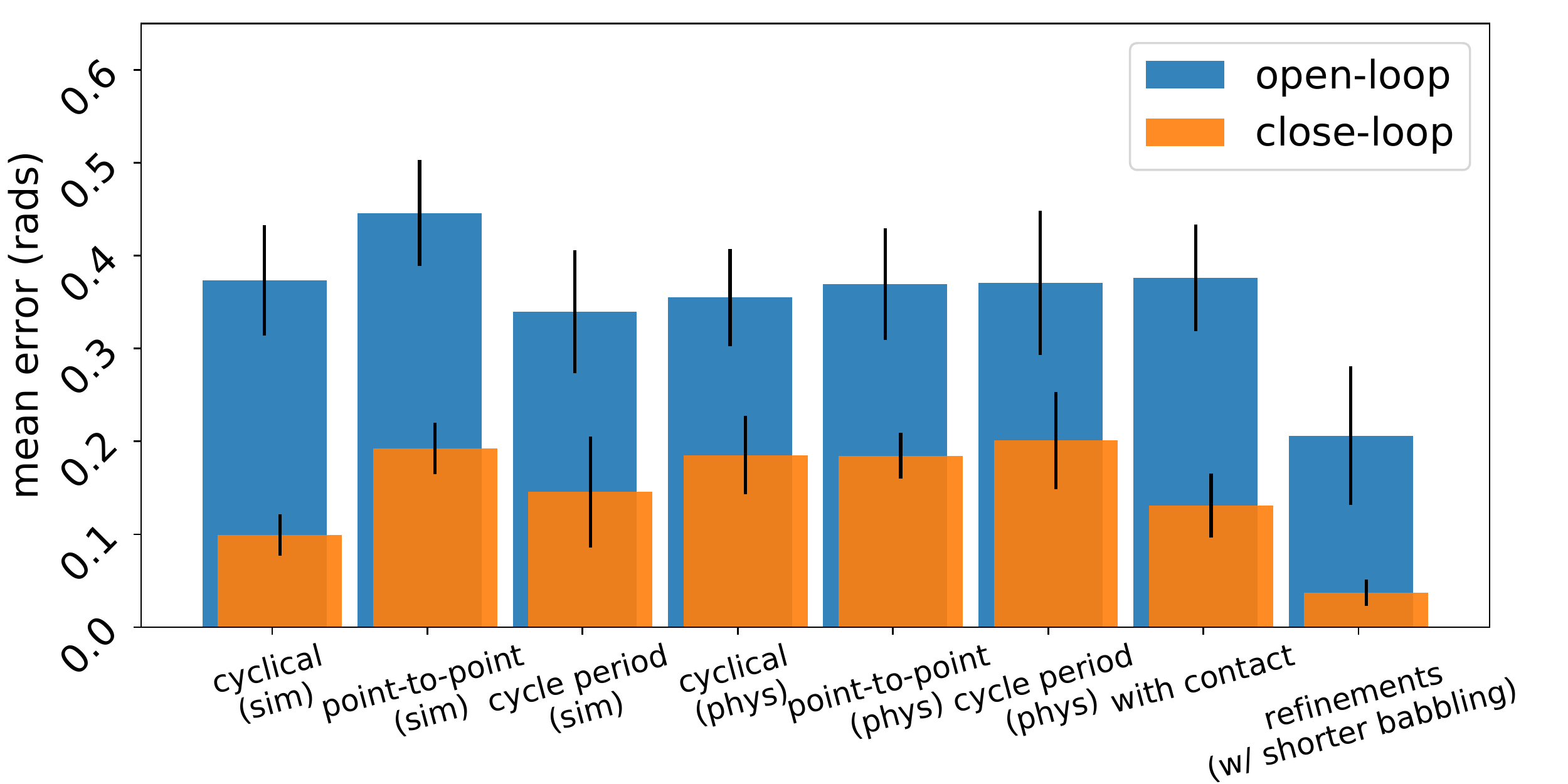}
\caption{The average error for open-loop and close-loop systems across all tasks.}
\label{fig:Fig3}
\end{figure}

In terms of the error (root mean square error between joint angles to the desired reference trajectories; will be referred to as ``error'' from here on), as expected, we see the close-loop control architecture reduces the error  compared to the open-loop one in all cases. Fig. \ref{fig:Fig3} shows the average error for the open-loop and the close-loop system across all tasks.

\subsection{Cyclical movements in-air task}
 
Fig. \ref{fig:Fig4} shows a sample trial of the cyclical movement in-air task for the physical system (also see supplementary video). This figure shows that the error is larger at the distal joint compared to the proximal joint. This is because all three tendons cross the proximal joint first, thus errors propagate to, and accumulate at, the distal joint.

For all ``sample run'' plots that were performed in both simulation and the physical system (Figs. \ref{fig:Fig4}, \ref{fig:Fig5}, and \ref{fig:Fig7}) we observe very similar patterns and are only reporting the physical system results here. The reader, however, can access all plots in \cite{marjaninejad2019simple}.

\begin{figure}
\includegraphics[width=\linewidth]{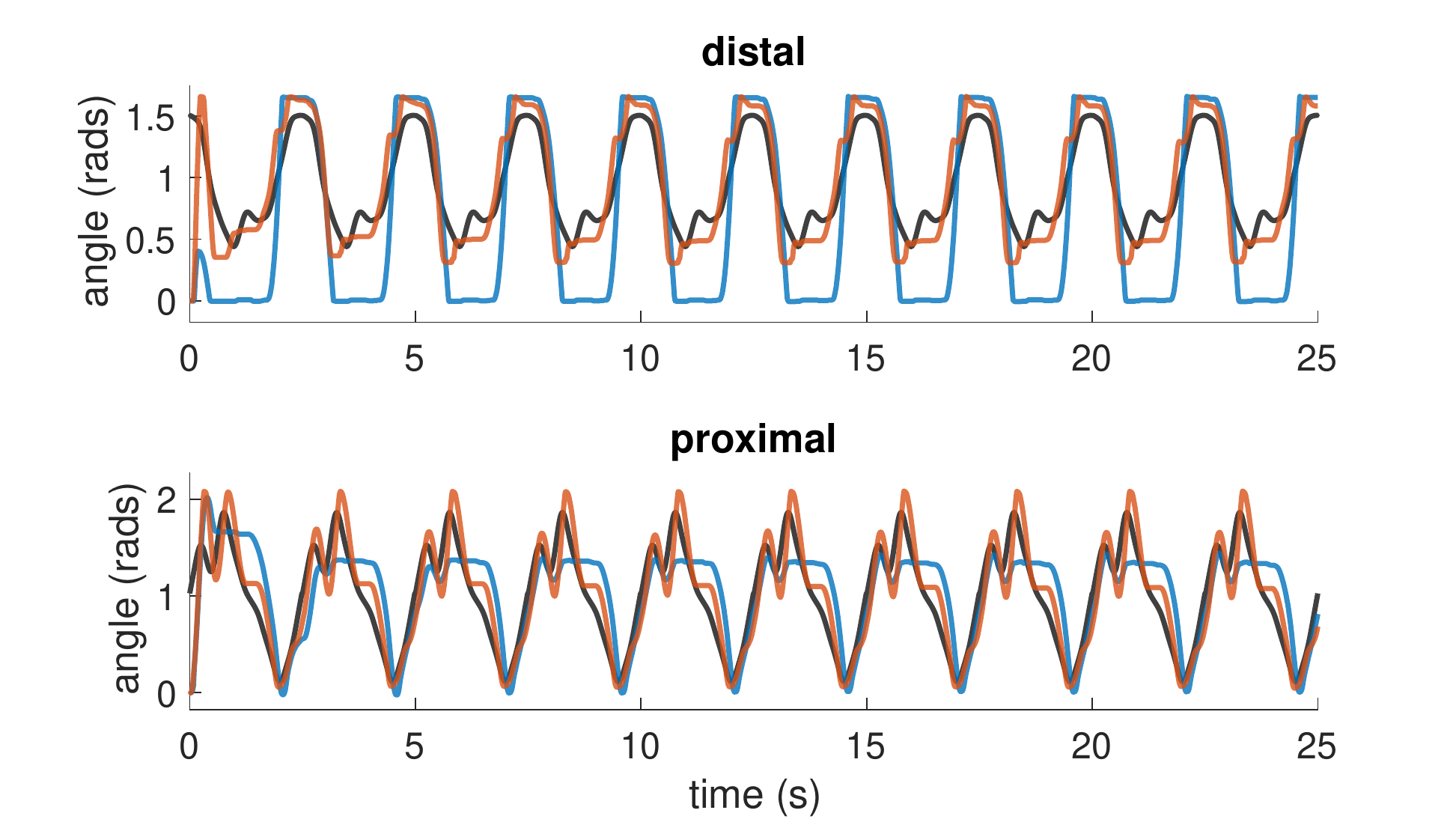}
\caption{The desired (black), open-loop (blue), and close-loop (orange) joint angles for one trial of the cyclical movements in-air task.}
\label{fig:Fig4}
\end{figure}

\subsection{Point-to-point movements in-air task}
\begin{figure}
\includegraphics[width=\linewidth]{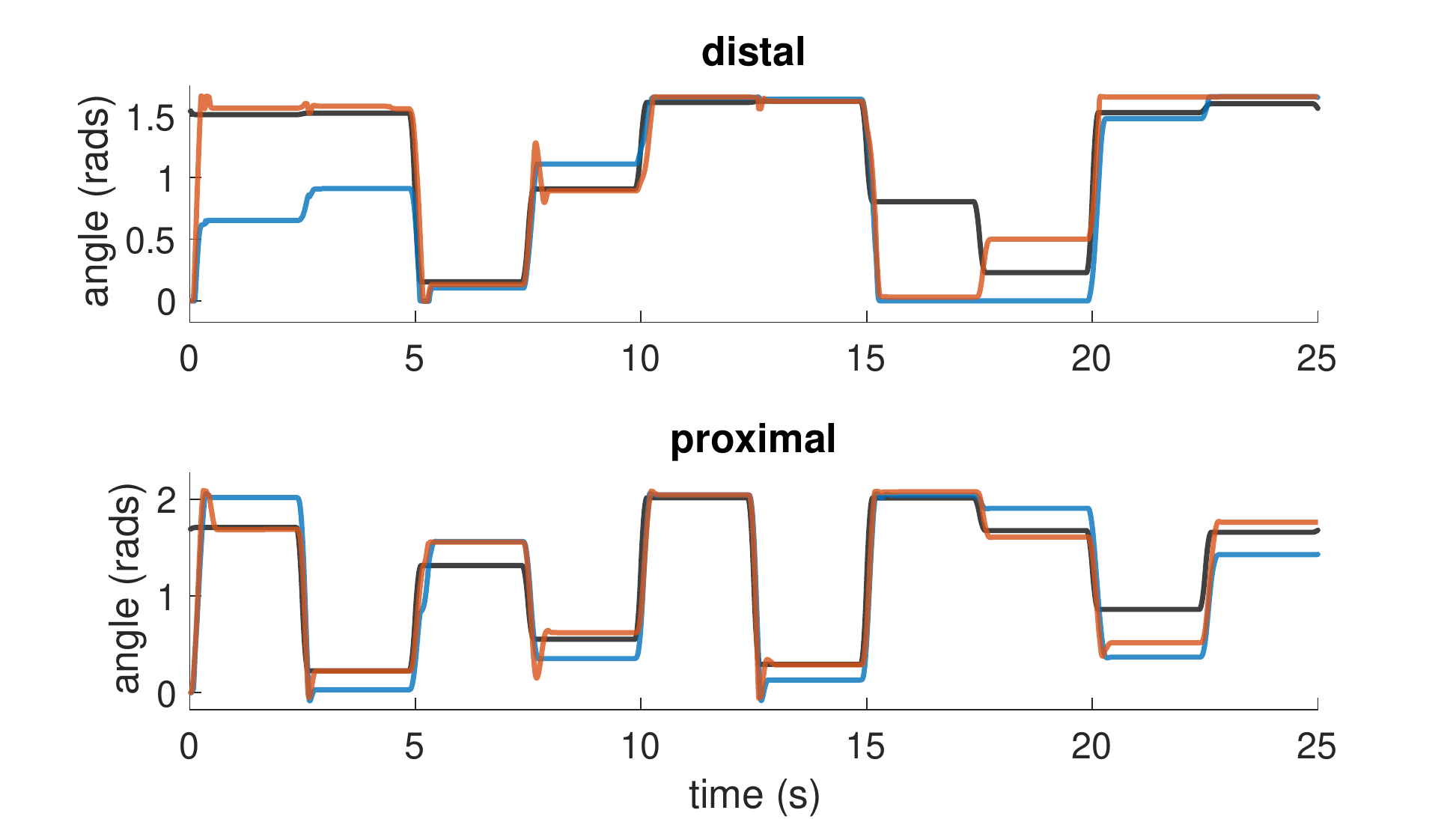}
\caption{The desired (black), open-loop (blue), and the close-loop (orange) joint angles for one trial of the point-to-point movements in-air task (over one sample run).}
\label{fig:Fig5}
\end{figure}

Figure~\ref{fig:Fig5} shows a good example of the limitations of an open-loop system (also see supplementary video). When the system is commanded to go to a new position, it can do so---except in cases where the commanded change is small. The inverse map may not have sufficient resolution to implement such small changes. Also, please note that both angular velocities and angular acceleration inputs will be zero (except during the transitions which are very short) and the system needs to go to the right position based only on joint angle values.  However, the close-loop system detects and corrects those errors. Importantly, this also improves the on-the-go training of the inverse map (see learning from each experience task). Note the unavoidable small fluctuations around the desired location, which are naturally caused by having a simple error correction feedback strategy.  Better tracking can be achieved with more sophisticated close-loop controllers, but that is beyond the objective and scope of this work.

\subsection{Different cycle period durations task}

\begin{figure}
\includegraphics[width=\linewidth]{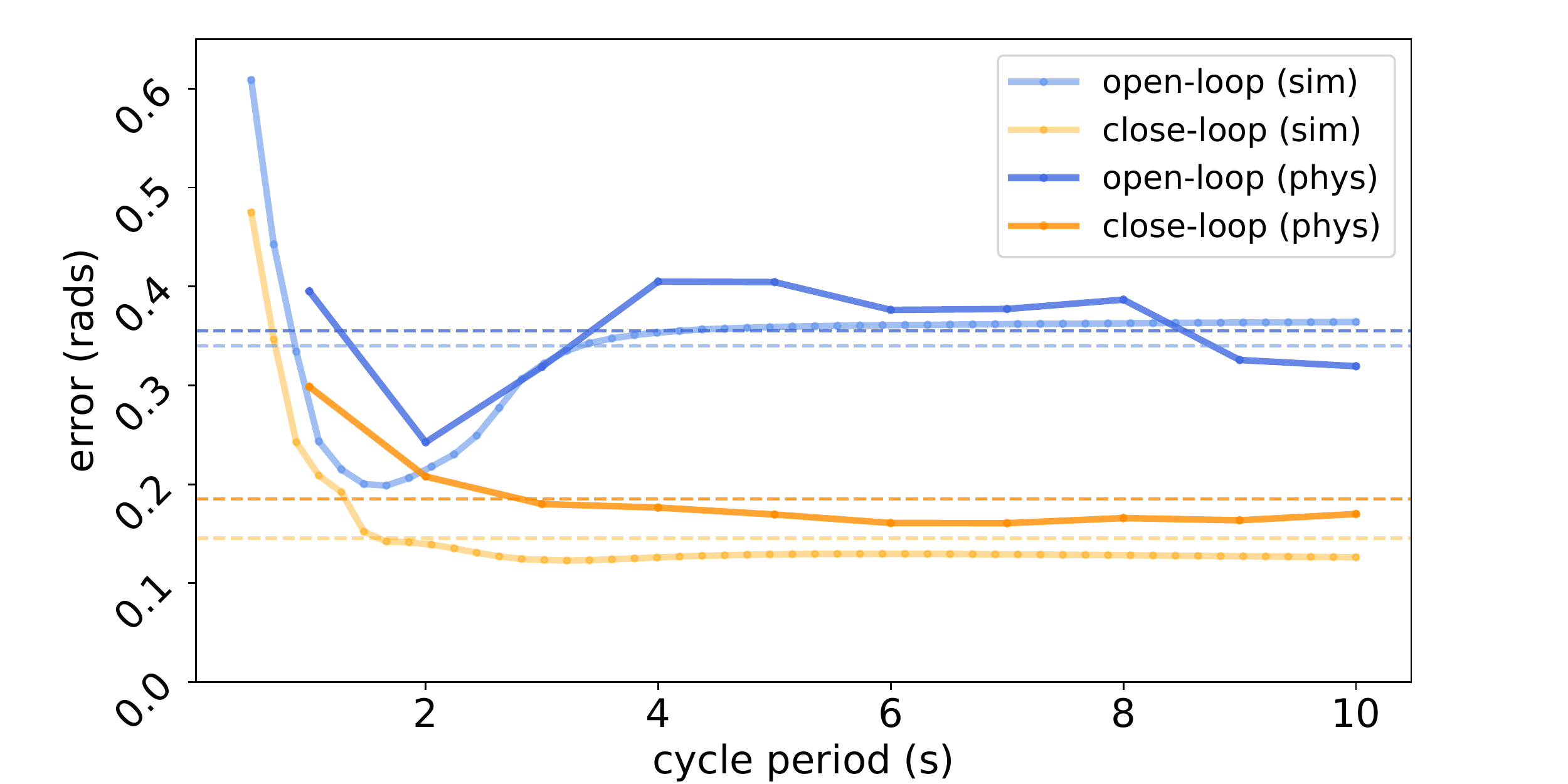}
\caption{Error values for different cycle period durations task as a function of cycle duration for the open-loop and the close-loop systems (sim: simulation; phys: physical system).}
\label{fig:Fig6}
\end{figure}

\begin{figure}
\includegraphics[width=\linewidth]{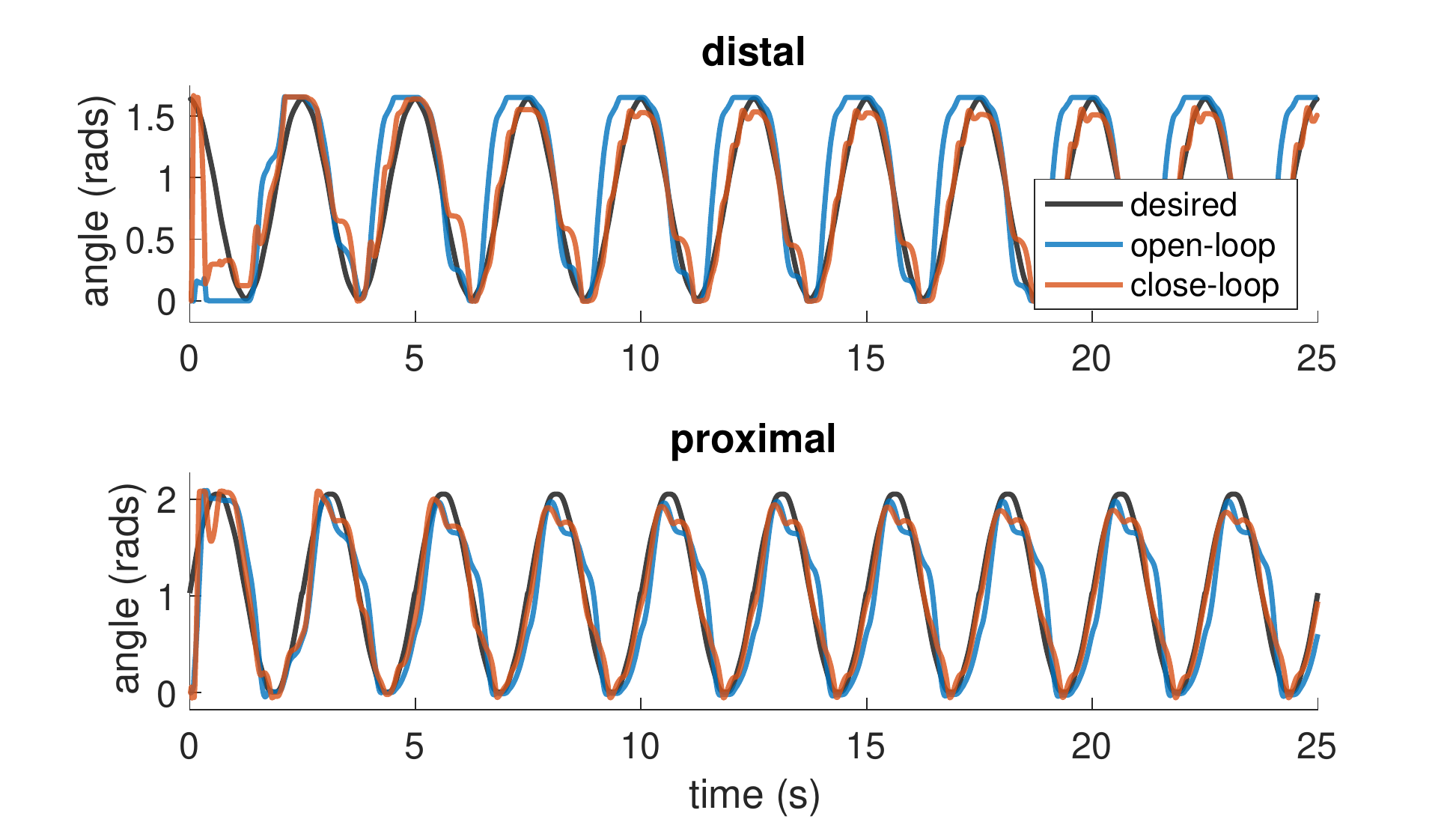}
\caption{The desired (black), open-loop (blue), and the close-loop (orange) joint angles for one sample run of the different cycle period durations task with a cycle period of 2.5 seconds.}
\label{fig:Fig7}
\end{figure}

The simple proportional-plus-integral feedback on joint angles has a limited bandwidth. We expect, and see, that its ability to correct errors degrades for cyclical movements with shorter cycle periods (i.e., higher frequencies).  Figure~\ref{fig:Fig6} shows improved performance of the close-loop system for cycle period durations longer than $\sim2$ seconds for both simulated and physical systems. As expected, the open-loop system also has problems at short cycle periods (due to effects of inertia, excitation of the nonlinearities of the double pendulum, bandwidth of the motors, etc), but the error does not improve as the cycle periods lengthen. The close-loop system  plateaus at a small average error of 0.1-0.2 rads per cycle quickly and then continues to reduce slowly. The error in the open-loop system, in contrast, has an error that is roughly twice as large with minimum at periods of $\sim1.5-2$ seconds, which is perhaps closer to the region it experienced during babbling and also close to the system's resonant frequency. Figure~\ref{fig:Fig7} shows the desired and actual outcomes of the task (for both open-loop and close-loop systems) over one sample run for a 2.5 seconds cycle period (also see supplementary video).

\subsection{Performance in the presence of contact dynamics tasks}
\subsubsection{Locomotion with the gantry}
This leg system was designed to ultimately produce locomotion. Therefore, we tested in simulation how the close- and open-loop systems performed this task. When introduced to mild ground contact (barely swiping the ground) both methods performed similarly well and comparably to the in-air task---albeit with a slightly larger error. However, when the simulated gantry was brought lower (and therefore more substantial contact dynamics were introduced), the open-loop system failed to clear the ground and could not complete the movement cycle to match the desired trajectories (see supplementary video and Figs. ~\ref{fig:Fig3} and ~\ref{fig:Fig8}). In contrast, the close-loop system was able to complete the swing-phase and recover from the ground contact (see Supplementary video), which then resulted in very small errors even in the presence of these contact dynamics.
This is expected as contact dynamics can be thought of as physical perturbations that were not included in the motor babbling. Thus the open-loop system naturally performs poorly (even with a well-refined, accurate inverse map). However, it was important to see that even simple feedback was able to compensate for such strong unmodeled, perturbations.

\begin{figure}
\includegraphics[width=\linewidth]{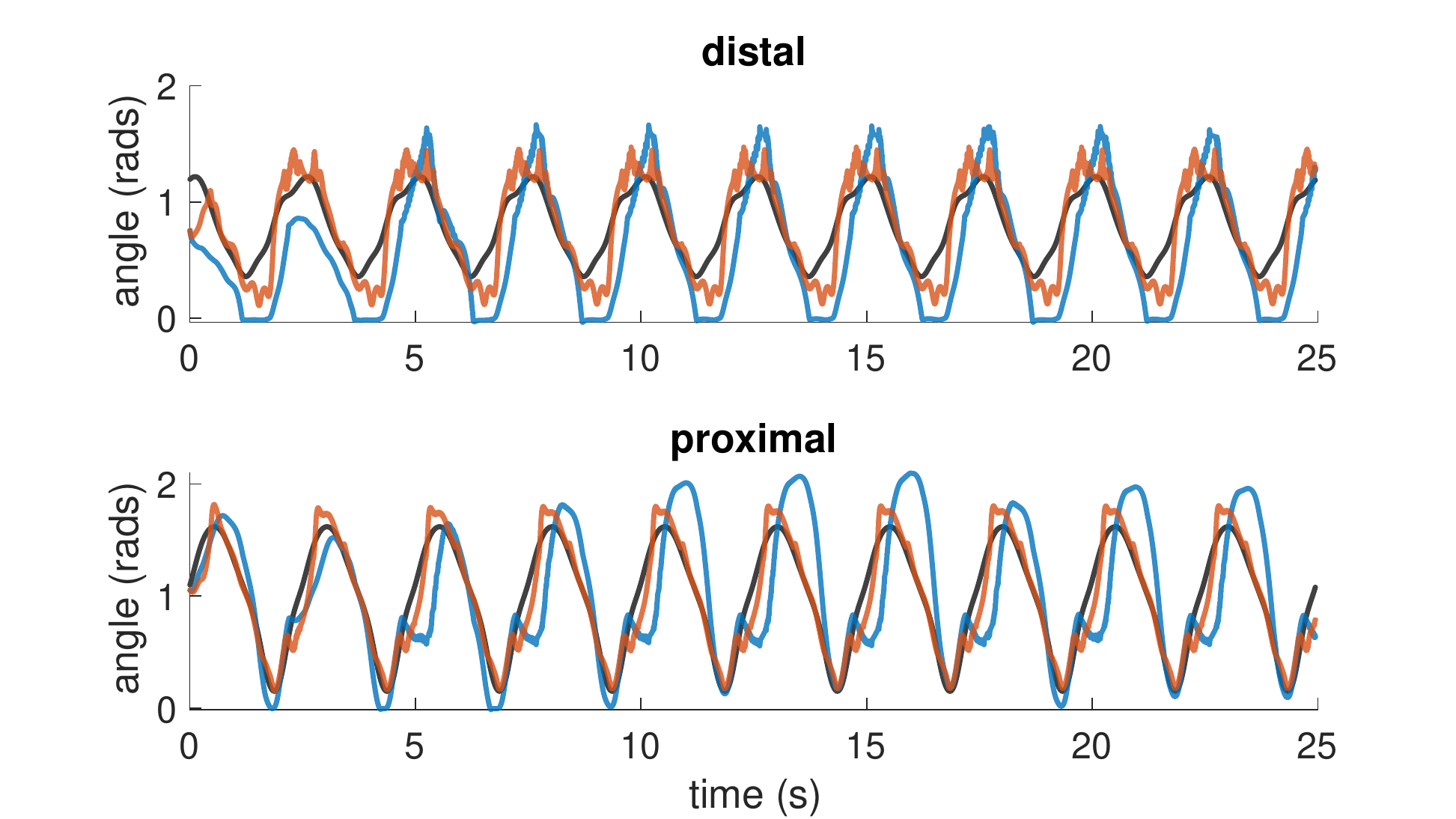}
\caption{The desired (black), open-loop (blue), and the close-loop (orange) joint angles for one sample run of the locomotion with the gantry task.}
\label{fig:Fig8}
\end{figure}

\subsubsection{Holding a posture under a weight}
In our simulations, we also observed that the open-loop system cannot compensate when a weight is applied to the chassis of the leg (the leg collapses under the weight). In contrast,   the close-loop system compensates---as much as the strengths of the muscles allow---for the deviation from the desired posture and maintains the prescribed posture which is standing vertically (see Supplementary Video).

\subsection{Learning from each experience task}
\begin{figure}
\includegraphics[width=.98\linewidth]{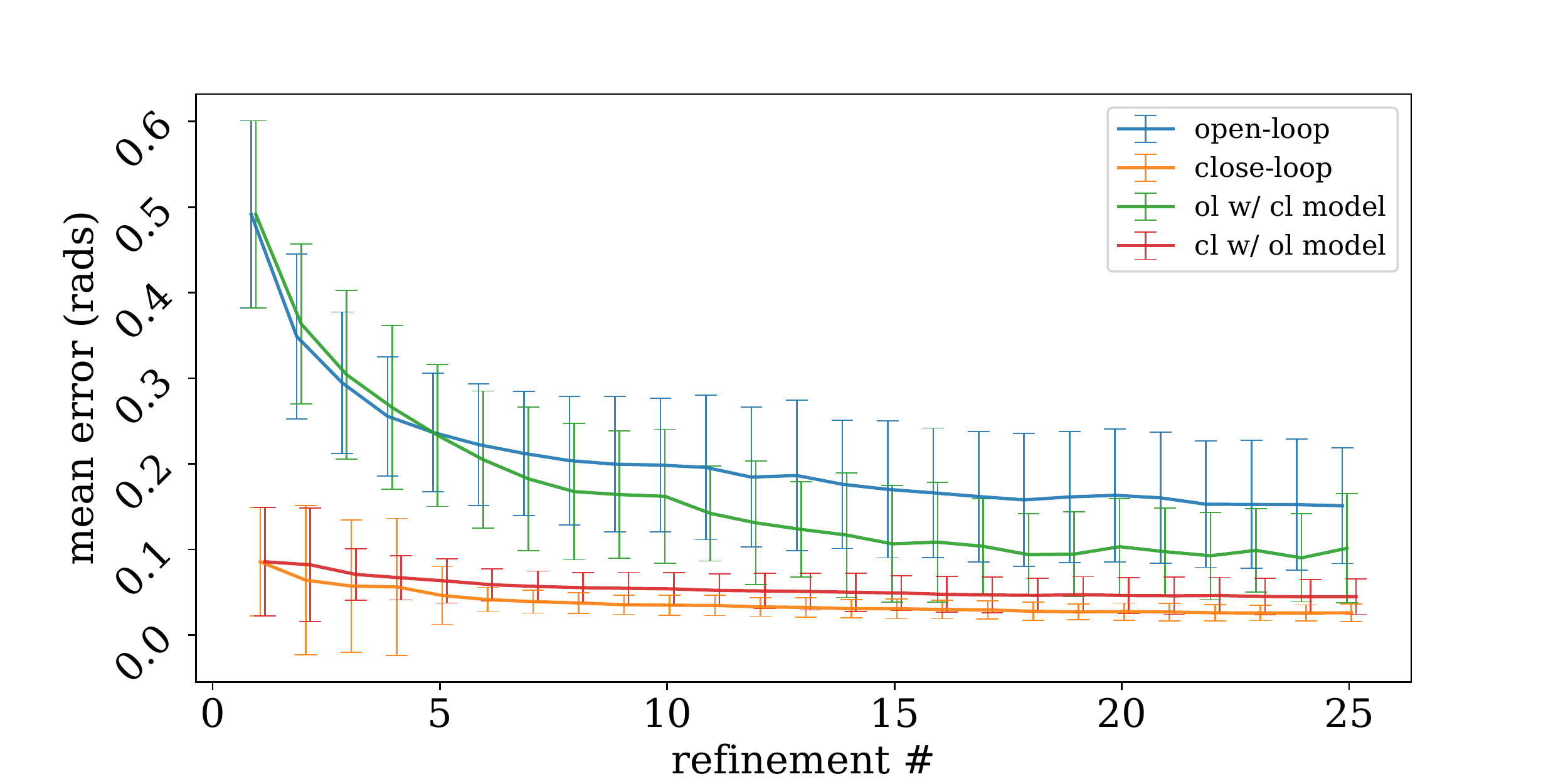}
\caption{Error values for 50 random cyclical movements as a function of repetitions (and consecutive refinements) for open-loop (blue) and close-loop (orange) systems as well as open-loop and close-loop with switched inverse maps (green and red, respectively).}
\label{fig:Fig9}
\end{figure}

Biological systems subject to Hebbian learning reinforce or attenuate synaptic connections with each experience~\cite{hebb1962organization, grillner2006biological}. Similarly, the G2P algorithm adds the input-output data from each run (i.e., experience) to its database and recalculates (i.e., refines) the inverse map with all available data before the next run (i.e., warm start of the ANN). Figure~\ref{fig:Fig9} shows the mean and standard deviation of error over 50 random cyclical movement tasks as a function of refinement number after each experience for both the open-loop (blue) and the close-loop (orange) systems. Both systems exhibit the expected reduction of error with increasing experience. However, this trend is accelerated in the close-loop system where both the mean and the standard deviation of the error plateau after only ~6 refinements.

We believe that the more relevant data collected by the close-loop system contributes to this. To test this idea, after each refinement, we tested both systems with switched inverse maps. This will distinguish contribution of the error correction of the feedback signal from the potential contribution of a more precise inverse map. Open-loop system shows accelerated learning and smaller error when using the inverse map trained by the close-loop data  (green). Also, although the error for the close-loop system with either inverse model is very small and plateaus fast (after $\sim5$ refinements), it has smaller mean and standard deviation with the inverse map trained with data collected by the close-loop system. The p-value between the 50 trials of the last refinement of close-loop systems using close-loop and open-loop inverse maps (orange and red curves, respectively) was 3.0927e-04. This measure for the open-loop systems (green and blue curves) was 1.0234e-07. These results show that not only does close-loop system reduce error by commanding correction signals, but it also enhances the refinements of the inverse map by providing more task specific data at each attempt.

To demonstrate that the system will not suffer from over-fitting with the proposed method and will allow generalization at the same time with improving the inverse map on-the-go, we also tested refinements for both systems for 50 random cyclical movements introduced back to back and saw similar descending trend in the error even for the movement cycles that were not experienced before. Results for simulation and physical system can be accessed from ~\cite{marjaninejad2019simple} and  ~\cite{marjaninejad2019autonomous},respectively. Note that the babbling data are always included in the refinements (as the original version of G2P ~\cite{marjaninejad2019autonomous} to make sure it will not over-fit to experience alone.

\subsection{Variable feedback delay task}
\begin{figure}
\includegraphics[width=\linewidth]{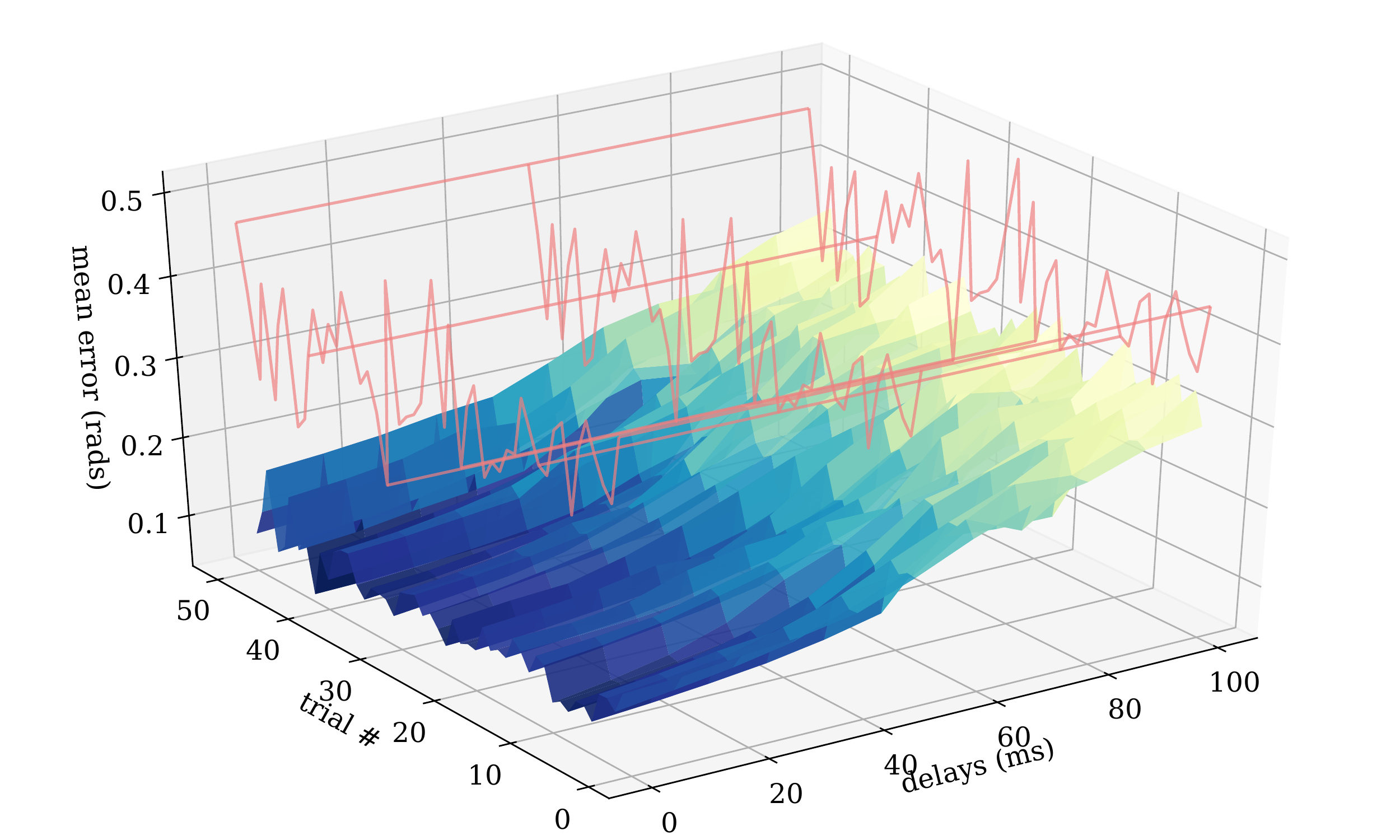}
\caption{Error values for 50 random cyclical movements as a function of feedback delay for the close-loop system. Error for the open-loop system is also provided (red wireframe styled lines) for comparison.}
\label{fig:Fig10}
\end{figure}

Figure~\ref{fig:Fig10} shows error over 50 random cyclical movements as a function of increasing feedback delay from 5--100 ms. We plot open-loop error (red wireframe styled lines) for the same tasks as a reference and see that the close-loop system outperforms it for delays up to 100 ms. At very long delays, naturally, the close-loop system will treat corrections as perturbations, and performance will degrade and instabilities will likely arise.

\subsection{Sensitivity to proportional-and-integral (PI) feedback gains}
The choice of PI gains is traditionally made by either trial-and-error, Bode plots and, more recently, by using search algorithms (e.g. evolutionary algorithms~\cite{geramipour2013design}). The choice of optimal PI gains is beyond the scope of this paper. However, we briefly explored the sensitivity to a wide range of PI gains over 50 cyclical movements and found that it still yields satisfactory performance (see Supplementary Information section in ~\cite{marjaninejad2019simple}---albeit with the expected faster rise times and greater overshoot with higher gains, and vice versa with lower gains.

\section{Summary of contributions}
Our method improves upon the current work in autonomous control of tendon-driven systems~\cite{marjaninejad2019autonomous} because i) it uses the inverse map of the tendon-driven system it autonomously learned during an initial motor babbling phase and only relies on feedback to compensate for inaccuracies as needed; and, more importantly, ii) shows that by collecting more relevant data during a performance, simple feedback also facilitates and accelerates autonomous learning; naturally, more relevant experience is more useful.

\section{Discussion}
We chose the input velocity signal to apply correction changes to as oppose to the input position signal since velocity has a direction (is a vector) and can move the joint to the right position even with an imperfect model (as can be seen from the point-to-point experiment results, the outputs of position input can have high errors). Also, please remember that we chose position as the output to define error on and it is very common in controls to use the derivative of the tracked signal for correction (Air conditioning, Cruise control, etc.). However, based on the need, used can choose any other error signal (or a weighted combination) and use PID gains to feed it to the most pertinent input to entertain the correction signals.

In this paper, we showed the contributions of simple kinematic feedback in improving both performance and learning rate of the inverse map generated using limited experience while being robust to sensory delays and choice of PI parameters. We performed our test in both simulation and physical implementations of a tendon-driven robotic limb. However, it would be very interesting to test the proposed system on more complex systems such as bipeds or quadrupeds and compare their performances in more sophisticated tasks in the future work, especially in their physical implementations.


%%%%%%%%%%%%%%%%%%%%%%%%%%%%%%%%%%%%%%%%%%%%%%%%%%%%%%%%%%%%%%%%%%%%%%%%%%%%%%%%

%%%%%%%%%%%%%%%%%%%%%%%%%%%%%%%%%%%%%%%%%%%%%%%%%%%%%%%%%%%%%%%%%%%%%%%%%%%%%%%%

%%%%%%%%%%%%%%%%%%%%%%%%%%%%%%%%%%%%%%%%%%%%%%%%%%%%%%%%%%%%%%%%%%%%%%%%%%%%%%%%
\section*{CODE AVAILABILITY}
The code and the supplementary files can be accessed through project's Github repository at: \url{https://github.com/marjanin/G2P_with_fb}

\section*{ACKNOWLEDGMENTS}
This project was supported by NIH Grants R01-052345 and R01-050520, award MR150091 by DoD, and award W911NF1820264 by DARPA-L2M program. Also, by USC Provost Fellowship to A.M. and the Consejo Nacional de Ciencia y Tecnología (Mexico) fellowship to D.U.-M.

%%%%%%%%%%%%%%%%%%%%%%%%%%%%%%%%%%%%%%%%%%%%%%%%%%%%%%%%%%%%%%%%%%%%%%%%%%%%%%%%

\bibliographystyle{IEEEtran}
\bibliography{IEEEabrv,Marjaninejad_bib}

\begin{thebibliography}{10}
\providecommand{\url}[1]{#1}
\csname url@rmstyle\endcsname
\providecommand{\newblock}{\relax}
\providecommand{\bibinfo}[2]{#2}
\providecommand\BIBentrySTDinterwordspacing{\spaceskip=0pt\relax}
\providecommand\BIBentryALTinterwordstretchfactor{4}
\providecommand\BIBentryALTinterwordspacing{\spaceskip=\fontdimen2\font plus
\BIBentryALTinterwordstretchfactor\fontdimen3\font minus
  \fontdimen4\font\relax}
\providecommand\BIBforeignlanguage[2]{{%
\expandafter\ifx\csname l@#1\endcsname\relax
\typeout{** WARNING: IEEEtran.bst: No hyphenation pattern has been}%
\typeout{** loaded for the language `#1'. Using the pattern for}%
\typeout{** the default language instead.}%
\else
\language=\csname l@#1\endcsname
\fi
#2}}

\bibitem{marjaninejad2019autonomous}
A.~Marjaninejad, D.~Urbina-Mel{\'e}ndez, B.~A. Cohn, and F.~J. Valero-Cuevas,
  ``Autonomous functional movements in a tendon-driven limb via limited
  experience,'' \emph{Nature Machine Intelligence}, vol.~1, no.~3, p. 144,
  2019.

\bibitem{kwiatkowski2019task}
R.~Kwiatkowski and H.~Lipson, ``Task-agnostic self-modeling machines,''
  \emph{Sci. Robot.}, vol.~4, p. eaau9354, 2019.

\bibitem{nagabandi2018neural}
A.~Nagabandi, G.~Kahn, R.~S. Fearing, and S.~Levine, ``Neural network dynamics
  for model-based deep reinforcement learning with model-free fine-tuning,'' in
  \emph{2018 IEEE International Conference on Robotics and Automation
  (ICRA)}.\hskip 1em plus 0.5em minus 0.4em\relax IEEE, 2018, pp. 7559--7566.

\bibitem{kobayashi2003adaptive}
H.~Kobayashi and R.~Ozawa, ``Adaptive neural network control of tendon-driven
  mechanisms with elastic tendons,'' \emph{Automatica}, vol.~39, no.~9, pp.
  1509--1519, 2003.

\bibitem{marques2014spontaneous}
H.~G. Marques, A.~Bharadwaj, and F.~Iida, ``From spontaneous motor activity to
  coordinated behaviour: a developmental model,'' \emph{PLoS computational
  biology}, vol.~10, no.~7, p. e1003653, 2014.

\bibitem{morimoto2001acquisition}
J.~Morimoto and K.~Doya, ``Acquisition of stand-up behavior by a real robot
  using hierarchical reinforcement learning,'' \emph{Robotics and Autonomous
  Systems}, vol.~36, no.~1, pp. 37--51, 2001.

\bibitem{gijsberts2013real}
A.~Gijsberts and G.~Metta, ``Real-time model learning using incremental sparse
  spectrum gaussian process regression,'' \emph{Neural Networks}, vol.~41, pp.
  59--69, 2013.

\bibitem{geijtenbeek2013flexible}
T.~Geijtenbeek, M.~Van De~Panne, and A.~F. Van Der~Stappen, ``Flexible
  muscle-based locomotion for bipedal creatures,'' \emph{ACM Transactions on
  Graphics (TOG)}, vol.~32, no.~6, p. 206, 2013.

\bibitem{rombokas2012tendon}
E.~Rombokas, E.~Theodorou, M.~Malhotra, E.~Todorov, and Y.~Matsuoka,
  ``Tendon-driven control of biomechanical and robotic systems: A path integral
  reinforcement learning approach,'' in \emph{2012 IEEE International
  Conference on Robotics and Automation}.\hskip 1em plus 0.5em minus
  0.4em\relax IEEE, 2012, pp. 208--214.

\bibitem{hunt2017development}
A.~Hunt, N.~Szczecinski, and R.~Quinn, ``Development and training of a neural
  controller for hind leg walking in a dog robot,'' \emph{Frontiers in
  neurorobotics}, vol.~11, p.~18, 2017.

\bibitem{kumar2014real}
V.~Kumar, Y.~Tassa, T.~Erez, and E.~Todorov, ``Real-time behaviour synthesis
  for dynamic hand-manipulation,'' in \emph{2014 IEEE International Conference
  on Robotics and Automation (ICRA)}.\hskip 1em plus 0.5em minus 0.4em\relax
  IEEE, 2014, pp. 6808--6815.

\bibitem{mnih2015human}
V.~Mnih, K.~Kavukcuoglu, D.~Silver, A.~A. Rusu, J.~Veness, M.~G. Bellemare,
  A.~Graves, M.~Riedmiller, A.~K. Fidjeland, G.~Ostrovski, \emph{et~al.},
  ``Human-level control through deep reinforcement learning,'' \emph{Nature},
  vol. 518, no. 7540, p. 529, 2015.

\bibitem{takahashi2017dynamic}
K.~Takahashi, T.~Ogata, J.~Nakanishi, G.~Cheng, and S.~Sugano, ``Dynamic motion
  learning for multi-dof flexible-joint robots using active--passive motor
  babbling through deep learning,'' \emph{Advanced Robotics}, vol.~31, no.~18,
  pp. 1002--1015, 2017.

\bibitem{heess2017emergence}
N.~Heess, S.~Sriram, J.~Lemmon, J.~Merel, G.~Wayne, Y.~Tassa, T.~Erez, Z.~Wang,
  S.~Eslami, M.~Riedmiller, \emph{et~al.}, ``Emergence of locomotion behaviours
  in rich environments,'' \emph{arXiv preprint arXiv:1707.02286}, 2017.

\bibitem{andrychowicz2018learning}
M.~Andrychowicz, B.~Baker, M.~Chociej, R.~Jozefowicz, B.~McGrew, J.~Pachocki,
  A.~Petron, M.~Plappert, G.~Powell, A.~Ray, \emph{et~al.}, ``Learning
  dexterous in-hand manipulation,'' \emph{arXiv preprint arXiv:1808.00177},
  2018.

\bibitem{schulman2017proximal}
J.~Schulman, F.~Wolski, P.~Dhariwal, A.~Radford, and O.~Klimov, ``Proximal
  policy optimization algorithms,'' \emph{arXiv preprint arXiv:1707.06347},
  2017.

\bibitem{schulman2015trust}
J.~Schulman, S.~Levine, P.~Abbeel, M.~Jordan, and P.~Moritz, ``Trust region
  policy optimization,'' in \emph{International conference on machine
  learning}, 2015, pp. 1889--1897.

\bibitem{bongard2006resilient}
J.~Bongard, V.~Zykov, and H.~Lipson, ``Resilient machines through continuous
  self-modeling,'' \emph{Science}, vol. 314, no. 5802, pp. 1118--1121, 2006.

\bibitem{valero2016fundamentals}
F.~J. Valero-Cuevas, \emph{Fundamentals of neuromechanics}.\hskip 1em plus
  0.5em minus 0.4em\relax Springer, 2016.

\bibitem{marjaninejad2019should}
A.~Marjaninejad and F.~J. Valero-Cuevas, ``Should anthropomorphic systems be
  “redundant”?'' in \emph{Biomechanics of Anthropomorphic Systems}.\hskip
  1em plus 0.5em minus 0.4em\relax Springer, 2019, pp. 7--34.

\bibitem{milton2009balancing}
J.~G. Milton, T.~Ohira, J.~L. Cabrera, R.~M. Fraiser, J.~B. Gyorffy, F.~K.
  Ruiz, M.~A. Strauss, E.~C. Balch, P.~J. Marin, and J.~L. Alexander,
  ``Balancing with vibration: a prelude for “drift and act” balance
  control,'' \emph{PLoS One}, vol.~4, no.~10, p. e7427, 2009.

\bibitem{cetinkaya2018stabilizing}
A.~Cetinkaya, T.~Hayakawa, and M.~A.~F. bin Mohd~Taib, ``Stabilizing unstable
  periodic orbits with delayed feedback control in act-and-wait fashion,''
  \emph{Systems \& Control Letters}, vol. 113, pp. 71--77, 2018.

\bibitem{cetinkaya2015sampled}
A.~Cetinkaya and T.~Hayakawa, ``Sampled-data delayed feedback control for
  stabilizing unstable periodic orbits,'' in \emph{2015 54th IEEE Conference on
  Decision and Control (CDC)}.\hskip 1em plus 0.5em minus 0.4em\relax IEEE,
  2015, pp. 1409--1414.

\bibitem{cohn2018feasibility}
B.~A. Cohn, M.~Szedl{\'a}k, B.~G{\"a}rtner, and F.~J. Valero-Cuevas,
  ``Feasibility theory reconciles and informs alternative approaches to
  neuromuscular control,'' \emph{Frontiers in computational neuroscience},
  vol.~12, 2018.

\bibitem{todorov2012mujoco}
E.~Todorov, T.~Erez, and Y.~Tassa, ``Mujoco: A physics engine for model-based
  control,'' in \emph{2012 IEEE/RSJ International Conference on Intelligent
  Robots and Systems}.\hskip 1em plus 0.5em minus 0.4em\relax IEEE, 2012, pp.
  5026--5033.

\bibitem{fazeli2019see}
N.~Fazeli, M.~Oller, J.~Wu, Z.~Wu, J.~Tenenbaum, and A.~Rodriguez, ``See, feel,
  act: Hierarchical learning for complex manipulation skills with multisensory
  fusion,'' \emph{Science Robotics}, vol.~4, no.~26, p. eaav3123, 2019.

\bibitem{fazeli2017learning}
N.~Fazeli, S.~Zapolsky, E.~Drumwright, and A.~Rodriguez, ``Learning
  data-efficient rigid-body contact models: Case study of planar impact,''
  \emph{arXiv preprint arXiv:1710.05947}, 2017.

\bibitem{marjaninejad2019simple}
A.~Marjaninejad, D.~Urbina-Mel{\'e}ndez, and F.~J. Valero-Cuevas, ``Simple
  kinematic feedback enhances autonomous learning in bio-inspired tendon-driven
  systems,'' \emph{arXiv preprint arXiv:1907.04539}, 2019.

\bibitem{hebb1962organization}
D.~O. Hebb, \emph{The organization of behavior: a neuropsychological
  theory}.\hskip 1em plus 0.5em minus 0.4em\relax Science Editions, 1962.

\bibitem{grillner2006biological}
S.~Grillner, ``Biological pattern generation: the cellular and computational
  logic of networks in motion,'' \emph{Neuron}, vol.~52, no.~5, pp. 751--766,
  2006.

\bibitem{geramipour2013design}
A.~Geramipour, M.~Khazaei, A.~Marjaninejad, and M.~Khazaei, ``Design of
  fpga-based digital pid controller using xilinx sysgen{\textregistered} for
  regulating blood glucose level of type-i diabetic patients,'' \emph{Int J
  Mechatron Electr Comput Technol}, vol.~3, no.~7, pp. 56--69, 2013.

\end{thebibliography}

\end{document}